\documentclass[a4paper, 10pt, conference]{ieeeconf}      
\usepackage{FG2019}
\usepackage{times}
\usepackage{epsfig}
\usepackage{graphicx}
\usepackage{amsmath}
\usepackage{amssymb}
\usepackage{multirow}
\usepackage{booktabs}
\usepackage{algorithm}
\usepackage[noend]{algpseudocode}
\usepackage{url}
\usepackage{subfigure}
\usepackage{calc}
\usepackage{amstext}
\usepackage{multicol}
\usepackage{pslatex}

\usepackage[singlelinecheck=false, small]{caption}
\newcommand{\RN}[1]{%
  \textup{\uppercase\expandafter{\romannumeral#1}}%
}

\FGfinalcopy 

\def\FGPaperID{76} 

\title{\LARGE \bf
Real-time Hand Gesture Detection and Classification Using Convolutional Neural Networks
}

\author{\parbox{16cm}{\centering
    {\large Okan K\"op\"ukl\"u$^1$, Ahmet Gunduz$^1$, Neslihan Kose$^2$, Gerhard Rigoll$^1$}\\
    {\normalsize
    \vspace{0.2cm}
    $^1$ Institute for Human-Machine Communication, TU Munich, Germany\\
    $^2$ Dependability Research Lab, Intel Labs Europe, Intel Deutschland GmbH, Germany}}
    \thanks{This work was not supported by any organization}
}

\begin{document}

\maketitle
\thispagestyle{plain}
\pagestyle{plain}


\begin{abstract}

Real-time recognition of dynamic hand gestures from video streams is a challenging task since (i) there is no indication when a gesture starts and ends in the video, (ii) performed gestures should only be recognized once, and (iii) the entire architecture should be designed considering the memory and power budget. In this work, we address these challenges by proposing a hierarchical structure enabling offline-working convolutional neural network (CNN) architectures to operate online efficiently by using sliding window approach. The proposed architecture consists of two models: (1) A detector which is a lightweight CNN architecture to detect gestures and (2) a classifier which is a deep CNN to classify the detected gestures. In order to evaluate the single-time activations of the detected gestures, we propose to use Levenshtein distance as an evaluation metric since it can measure misclassifications, multiple detections, and missing detections at the same time. We evaluate our architecture on two publicly available datasets - EgoGesture and NVIDIA Dynamic Hand Gesture Datasets - which require temporal detection and classification of the performed hand gestures. ResNeXt-101 model, which is used as a classifier, achieves the state-of-the-art offline classification accuracy of 94.04\% and 83.82\% for depth modality on EgoGesture and NVIDIA benchmarks, respectively. In real-time detection and classification, we obtain considerable early detections while achieving performances close to offline operation. The codes and pretrained models used in this work are publicly available \footnote{https://github.com/ahmetgunduz/Real-time-GesRec}.

\end{abstract}
\section{Introduction}

Computers and computing devices are becoming an essential part of our lives day by day. The increasing demand for such computing devices increased the necessity of easy and practical computer interfaces. For this reason, systems using vision-based interaction and control are becoming more common, and as a result of this, gesture recognition is getting more and more popular in research community due to various application possibilities in human machine interaction. Compared to mouse and keyboard, any vision-based interface is more convenient, practical and  natural because of the intuitiveness of gestures. 


\begin{figure}[t!]
	\centering
	\includegraphics[width=0.52\textwidth]{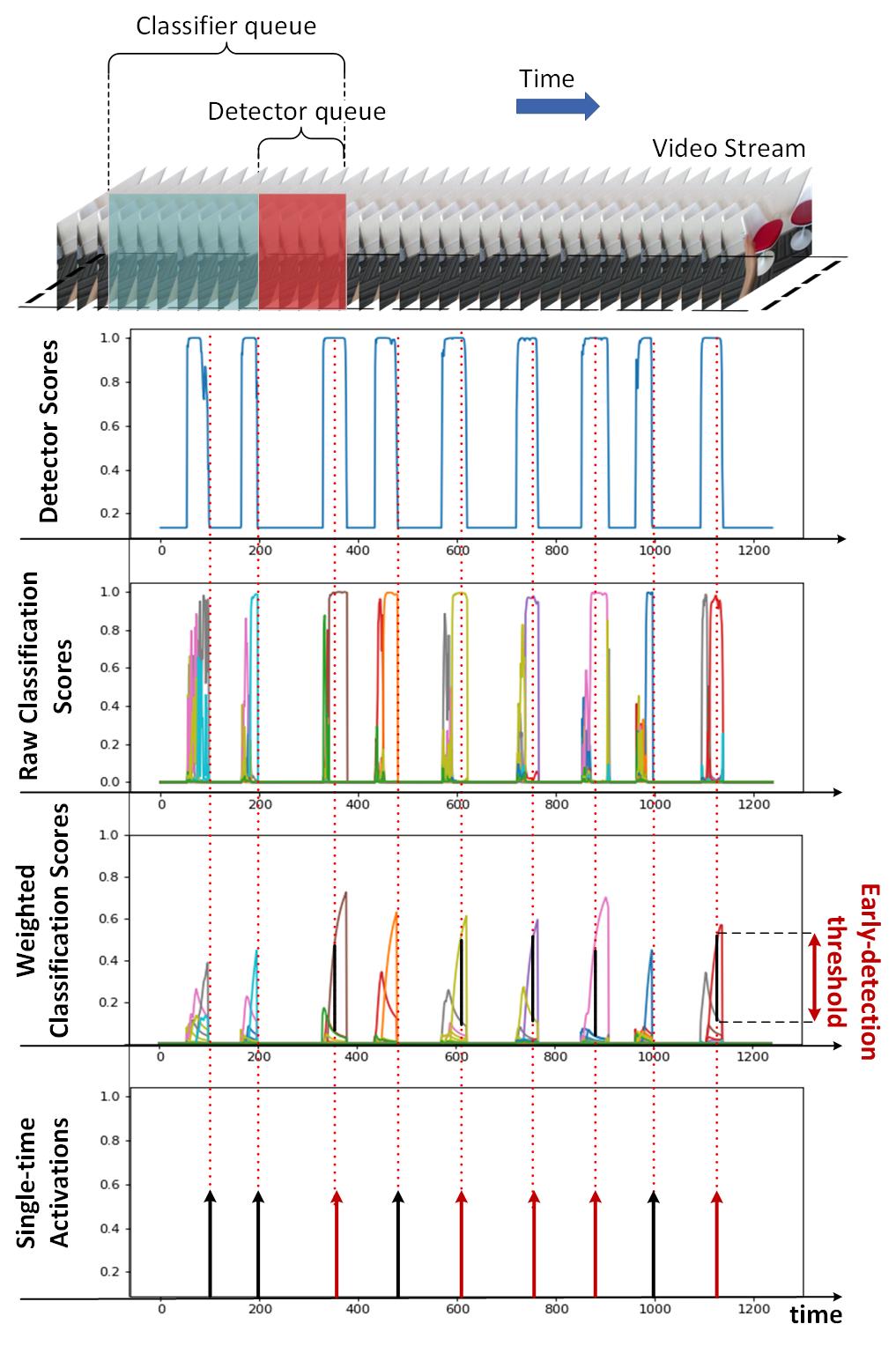}
	\caption{Illustration of the proposed pipeline for real-time gesture recognition. The video stream is processed using a sliding window approach with stride of one. Top graph shows the detector probability scores which is activated when a gesture starts and kept active till it ends. The second graph shows classification score for each class with a different color. The third graph applies weighted-average filtering on raw classification scores which eliminates the ambiguity between possible gesture candidates. The bottom graph illustrates the single-time activations such that red arrows represent early detections and black ones represent detections after gestures finalize.}
	\label{fig:probs}
\end{figure}

Gesture recognition can be practiced with mainly three methods: Using \textit{(i)} glove-based wearable devices \cite{abhishek_glove-based_2016}, \textit{(ii)} 3-dimensional locations of hand keypoints \cite{wen_intraoperative_2010} and \textit{(iii)} raw visual data. The first method comes with the obligation of wearing an additional device with which lots of cables come even though it provides good results in terms of both accuracy and speed. The second, on the other hand, requires an extra step of hand-keypoints extraction, which brings additional time and computational cost. Lastly, for (iii), only an image capturing sensor is required such as camera, infrared sensor or depth sensor, which are independent of the user. Since the user does not require to wear a burdensome device to achieve an acceptable accuracy in recognition and sufficient speed in computation, this option stands out as the most practical one. It is important for the infrastructure of any gesture recognition system to be practical. After all, we aim to use it in real life scenarios. 

In this work, in order to provide a practical solution, we have developed a vision based gesture recognition approach using deep convolutional neural networks (CNNs) on raw video data. Currently, CNNs provide the state-of-the-art results for not only image based tasks such as object detection, image segmentation and classification, but also for video based tasks such as activity recognition and action localization as well as gesture recognition \cite{kopuklu2018motion, molchanov2016online, zhu2017multimodal}.  

In real-time gesture recognition applications, there are several characteristics that the system needs to satisfy: \textit{(i)} An acceptable classification accuracy, \textit{(ii)} fast reaction time, \textit{(iii)} resource efficiency and \textit{(iv)} single-time activation per each performed gesture. All these items contain utmost importance for a successful real-time vision based gesture recognition application. However, most of the previous research only considers \textit{(i)} and tries to increase the offline classification accuracy in gesture recognition disregarding the remaining items. Some proposed approaches are even impossible to run in real-time since they consist of several deep CNNs on multiple input modalities, which is forcing the limits of memory and power budget \cite{narayana2018gesture}. 

In this paper, we propose a hierarchical architecture for the task of real-time hand gesture detection and classification that allows us to integrate offline working models and still satisfy all the above-mentioned attributes. Our system consists of an offline-trained deep 3D CNN for gesture classification (classifier) and a light weight, shallow 3D CNN for gesture detection (detector). Fig. \ref{fig:probs} illustrates the pipeline of the proposed approach. A sliding window is used over the incoming video stream feeding the input frames to the detector via detector queue. Top graph in Fig. \ref{fig:probs} shows the detector probability scores which become active when the gestures are being performed, and remain inactive for the rest of the time. The classifier becomes active only when the detector detects a gesture. This is very critical since most of the time, no gesture is performed in real-time gesture recognition applications. Therefore, there is no need to keep the high-performance classifier always active, which increases the memory and power consumption of the system considerably. The second graph shows the raw classification scores of each class with a different color. As it can be seen from the graph, scores of similar classes become simultaneously high especially at the beginning of the gestures. In order to resolve these ambiguities, we have weighted the class scores to avoid making a decision at the beginning of the gestures (third graph in Fig. \ref{fig:probs}). Lastly, the bottom graph illustrates the single-time activations, where red arrows represent the early detections and black ones represent the detections after gestures end. Our system can detect gestures earlier in their nucleus part, which is the part distinguishing the gesture from the rest. We propose to use the Levenshtein distance as an evaluation metric to compare the captured single-time activations with ground-truth labels. This metric is more suitable and evaluative since it can measure misclassifications, multiple detections and missing detections at the same time. 

We evaluated our approach on two publicly available datasets, which are EgoGesture Dataset \cite{zhang_egogesture:_2018} and NVIDIA Dynamic Hand Gesture Dataset \cite{molchanov2016online} (nvGesture) \footnote{NVIDIA Dynamic Hand Gesture Dataset is referred as 'nvGesture' in this work.}. For the classifier of the proposed approach, any offline working CNN architecture can be used. For our experiments, we have used well-known C3D \cite{tran2015learning} and ResNeXt-101 \cite{hara3dcnns}. We have achieved the state-of-the-art offline classification accuracies of 94.03\% and 83.82\% on depth modality with ResNeXt-101 architecture on EgoGesture and nvGesture datasets, respectively. For real-time detection and classification, we achieve considerable early detections by relinquishing little amount of recognition performance.

The rest of the paper is organized as follows. In Section \RN{2}, the related work in the area of offline and real-time gesture recognition is presented. Section \RN{3} introduces our real-time gesture recognition approach, and elaborates training and evaluation processes. Section \RN{4} presents experiments and results. Lastly, Section \RN{5} concludes the paper.
\section{Related Work}

The success of CNNs in object detection and classification tasks \cite{krizhevsky2012imagenet, Girshick2014rich} has created a growing trend to apply them also in the other areas of computer vision. For video analysis tasks, CNNs have been initially extended to be applied for video action and activity recognition and they have achieved state-of-the-art performances \cite{simonyan2014two, donahue2015long}.

There have been various approaches using CNNs to extract spatio-temporal information from video data. Due to the success of 2D CNNs in static images, video analysis based approaches initially applied 2D CNNs. In \cite{simonyan2014two, Karpathy2014largescale}, video frames are treated as multi-channel inputs to 2D CNNs. Temporal Segment Network (TSN) \cite{wang2016temporal} divides video into several segments, extracts information from color and optical flow modalities for each segment using 2D CNNs, and then applies spatio-temporal modeling for action recognition. A convolutional long short-term memory (LSTM) architecture is proposed in \cite{donahue2015long}, where the authors extract first the features from video frames by a 2D CNN and then apply LSTM for global temporal modeling. The strength of all these approaches comes from the fact that there are plenty of very successful 2D CNN architectures, and these architectures can be pretrained using the very large-scale ImageNet dataset \cite{deng2009imagenet}.

Although 2D CNNs perform pretty well on video analysis tasks, they are limited to model temporal information and motion patterns. Therefore 3D CNNs have been proposed in \cite{tran2015learning, tran2017convnet, hara3dcnns}, which use 3D convolutions and 3D pooling to capture discriminative features along both spatial and temporal dimensions. Different from 2D CNNs, 3D CNNs take a sequence of video frames as inputs. In this work, we also use the variants of 3D CNNs.


The real-time systems for hand gesture recognition requires to apply detection and classification simultaneously on continuous stream of video. There are several works addressing detection and classification separately. In \cite{ohn2014hand}, authors apply histogram of oriented gradient (HOG) algorithm together with an SVM classifier. The authors in \cite{molchanov2015multi} use a special radar system to detect and segment gestures. In our work, we have trained a light weight 3D CNN for gesture detection. Moreover, in human computer interfaces, performed gestures must be recognized only once (i.e. single-time activations) by the computers. This is very critical and this problem has not been addressed well yet. In \cite{molchanov2016online}, the authors apply connectionless temporal classification (CTC) loss to detect consecutive similar gestures. However, CTC does not provide single time activations. To the best of our knowledge, in this study, it is the first time single-time activations are performed for deep learning based hand gesture recognition.   

\section{Methodology}

In this section, we elaborate on our two-model hierarchical architecture that enables the-state-of-the-art CNN models to be used in real-time gesture recognition applications as efficiently as possible. After introducing the  architecture, training details are described. Finally, we give a detailed explanation for the used post processing strategies that allow us to have single-time activation per gesture in real-time.  

\subsection{Architecture}

Recently, with the availability of large datasets, CNN based models have proven their ability in action/gesture recognition tasks. 3D CNN architectures especially stand out for video analysis since they make use of the temporal relations between frames together with their spatial content. However, there is no clear description of how to use these models in a real-time dynamic system. With our work, we aim to fill this research gap.

\begin{figure}[t!]
	\centering
	\includegraphics[width = 0.50\textwidth]{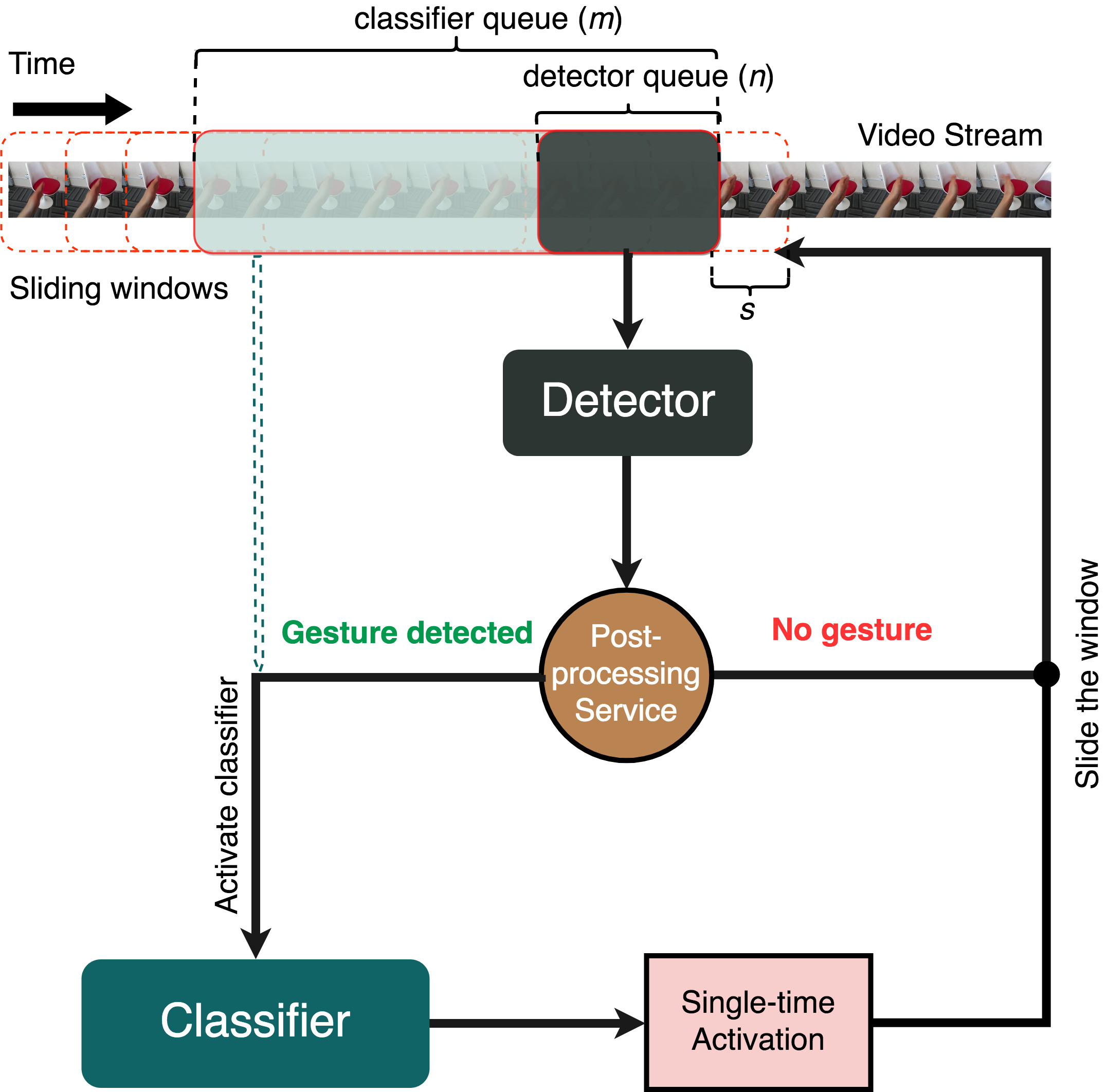}
	\caption{The general workflow of the proposed two-model hierarchical architecture. Sliding windows with stride \textit{s} run through incoming video frames where detector queue placed at the very beginning of classifier queue. If the detector recognizes an action/gesture, then the classifier is activated. The detector's output is post-processed for a more robust performance, and the final decision is made using single-time activation block where only one activation occurs per performed gesture.}
	\label{fig:workflow}
\end{figure}

Fig. \ref{fig:workflow} illustrates the used workflow for an efficient real-time recognition system using a sliding window approach. In contrary to offline testing, we do not know when a gesture starts or ends. Because of this, our workflow starts with a detector which is used as a switch to activate classifier if a gesture gets detected. Our detector and classifier models are fed by a sequence of frames with size $n$ and $m$, respectively, such as $n \ll m$ with an overlapping factor as shown in Fig. \ref{fig:workflow}. The stride value used for the sliding window is represented by $s$ in Fig. \ref{fig:workflow}, and it is same for both the detector and the classifier. Although higher stride provides less resource usage, we have chosen $s$ as 1 since it is small enough not to miss any gestures and allows us to achieve better performance. In addition to the detector and classifier models, one post-processing and one single-time activation service is introduced to the workflow. In the following parts, we are going to explain these blocks in detail.

\subsubsection{Detector}

The purpose of the detector is to distinguish between \textit{gesture} and \textit{no gesture} classes by running on a sequence of images, which detector queue masks. Its main and only role is to act as a switch for the classifier model, meaning that if it detects a \textit{gesture}, then the classifier is activated and fed by the frames in the classifier queue.

\begin{figure}[t!]
	\centering
	\includegraphics[width=0.30\textwidth]{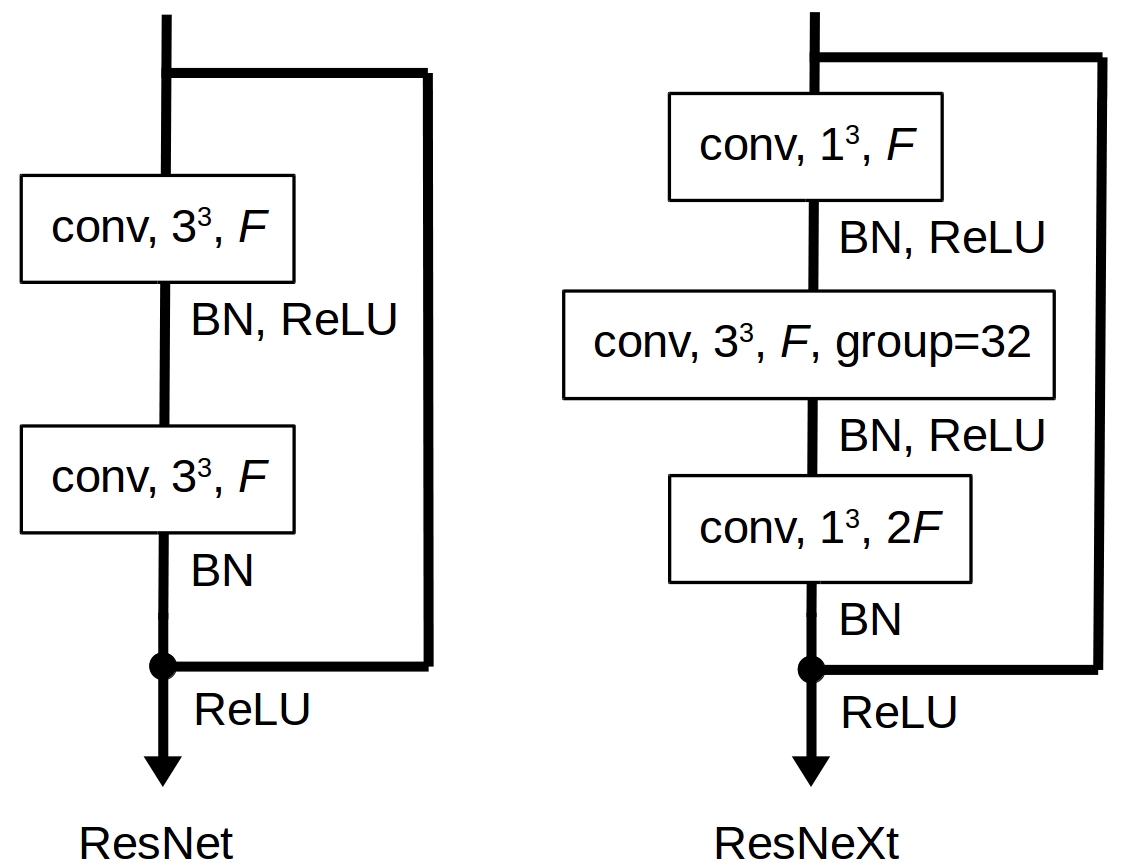}
	\caption{ResNet and ResNeXt blocks used in the detector and classifier architectures.}
	\label{fig:blocks}
\end{figure}

Since the overall accuracy of this system highly depends on the performance of detector, we require the detector to be (i) robust, (ii) accurate in detection of true positives (gestures), and (iii) lightweight as it runs continuously. For the sake of (i), the detector runs on smaller number of frames than the classifier to which we refer as detector and classifier queues. For (ii), detector queue is placed on the very beginning of classifier queue as shown in Fig. \ref{fig:workflow}, and this enables the detector to activate the classifier whenever a gesture starts regardless of the gesture duration. Moreover, the detector model is trained with a weighted-cross entropy loss in order to decrease the likelihood of false positives (i.e., achieve higher recall rate). The class weights for \textit{no gesture} and \textit{gesture} classes are selected as 1 and 3, respectively as our experiments showed that this proportion is sufficient to have 98+\% and 97+\% recall rates in EgoGesture and nvGesture datasets, respectively. Besides that, we post-process the output probabilities, and set a counter for the consecutive number of \textit{no gesture} predictions in decision of deactivating classifier. For (iii), ResNet-10 architecture is constructed using the ResNet block in Fig. \ref{fig:blocks} with very small feature sizes in each layer as given in Table \ref{tab:architecture}, which results in less than 1M ($\approx$ 862K) parameters. \textit{F} and \textit{N} correspond to the number of feature channels and the number blocks in corresponding layers, respectively. \textit{BN}, \textit{ReLU} and \textit{group} in Fig. \ref{fig:blocks} refers to batch normalization, rectified linear unit nonlinearities and the number of group convolutions, respectively.

\subsubsection{Classifier}

Since we do not have any limitation regarding the size or complexity of the model, any architecture providing a good classification performance can be selected as classifier. This leads us to use two recent 3D CNN architectures (C3D \cite{tran_learning_2014} and ResNext-101 \cite{he_deep_2015}) as our classifier model.  However, it is important to note that our architecture is independent of the model type.

For C3D model, we have used the exact same model as in \cite{tran_learning_2014}, but only changed the number of nodes in the last two fully connected layers from 4096 to 2048. For ResNeXt-101, we have followed the guidelines of \cite{hara3dcnns} and chosen the model parameters as given in Table \ref{tab:architecture} with ResNeXt block as given in Fig. \ref{fig:blocks}.

Since the number of parameters for 3D CNNs are much more than 2D CNNs, they require more training data in order to prevent overfitting. Because of this reason, we pretrained our classifier architectures first on Jester dataset \cite{jester}, which is the largest publicly available hand gesture dataset, and then fine tune our model on EgoGesture and nvGesture datasets. This approach increased the accuracy and shortened the training duration drastically.

\textit{Training Details: } We use stochastic gradient descent (SGD) with Nesterov momentum = 0.9, damping factor = 0.9, and weight decay = 0.001 as optimizer. After pretraining on Jester dataset, the learning rate is started with 0.01, and divided by 10 at $10^{th}$ and $25^{th}$ epochs, and training is completed after 5 more epochs.

For regularization, we used a weight decay ($\gamma = 1 \times 10^{-3}$), which is applied on all the parameters of the network. We also used dropout layers in C3D and several data augmentation techniques throughout training.

For data augmentation, three methods were used: (1) Each image is randomly cropped with size $112 \times 112$ and scaled randomly with one of \{1, $\frac{1}{2^{1/4}}$, $\frac{1}{2^{3/4}}$, $\frac{1}{2}$\} scales. (2) Spatial elastic displacement \cite{simard2003best} with $\alpha = 1$ and $\sigma = 2$ is applied on the cropped and scaled images. For temporal augmentation, (3) we randomly select consecutive frames according to the size of input sample duration from the entire gesture videos. If the sample duration is more than the number of frames in target gesture, we append frames starting from the very first frame in a cyclic fashion. We also normalized the images into 0-1 scale using mean and standard deviation of the whole training sets in order to force models to learn faster. The same training details are used for the detector and classifier models.

During offline and online testing, we scale images and apply center cropping to get $112 \times 112$ images. Then only normalization is performed for the sake of consistency between training and testing. 

\begin{table}[t!]
	\centering
	\begin{tabular}{l|c|c|c}
		\hline \rule{0pt}{12pt}
		\textbf{Layer}    & \textbf{Output Size} & \textbf{ResNeXt-101}  & \textbf{ResNet-10} \\[0.1cm] \hline \rule{0pt}{12pt} 
		conv1    &  L x 56 x 56 & \multicolumn{2}{c}{conv(3x7x7), stride (1, 2, 2)}  \\[0.1cm] \hline \rule{0pt}{12pt}
		pool    &  L/2 x 28 x 28 & \multicolumn{2}{c}{ MaxPool(3x3x3), stride (2, 2, 2)}  \\[0.1cm] \hline \rule{0pt}{12pt}
		conv2\_x &  L/2 x 28 x 28 & N:3, F:128   & N:1, F:16  \\[0.1cm] \hline \rule{0pt}{12pt} 
		conv3\_x &  L/4 x 14 x 14 & N:24, F:256  & N:1, F:32  \\[0.1cm] \hline \rule{0pt}{12pt}
		conv4\_x &  L/8 x 7 x 7 & N:36, F:512  & N:1, F:64  \\[0.1cm] \hline \rule{0pt}{12pt}
		conv5\_x &  L/16 x 4 x 4 & N:3, F:1024  & N:1, F:128 \\[0.1cm] \hline \rule{0pt}{14pt}
		         & \textit{NumCls} & \multicolumn{2}{c}{\begin{tabular}[c]{@{}c@{}}global average pooling,\\ fc layer with softmax\end{tabular}} \\[0.2cm] \hline
	\end{tabular}
	\caption{Detector (ResNet-10) and Classifier (ResNeXt-101) architectures. For ResNet-10, max pooling is not applied when input of 8-frames is used.}
	\label{tab:architecture}
\end{table}

\subsubsection{Post-processing}
\label{subsec:pp}

In dynamic hand gestures, it is possible that the hand gets out of the camera view while performing gestures. Even though the previous predictions of the detector are correct, any misclassification reduces the overall performance of the proposed architecture. In order to make use of previous predictions, we add the raw softmax probabilities of the previous detector predictions into a queue ($q_k$) with size $k$, and apply filtering on these raw values and obtain final detector decisions. With this approach, detector increases its confidence in decision making, and clears out most of the misclassifications in consecutive predictions. The size of the queue ($k$) is selected as 4, which achieved the best results for stride $s$ of 1 in our experiments. 

We have applied $(i)$ average, $(ii)$ exponentially-weighted average and $(iii)$ median filtering separately on the values in $q_k$. While average filtering simply takes the mean value of $q_k$, median filtering takes the median. Exponentially-weighted average filtering, on the other hand, takes the weighted average of the samples using the weight function of  $w_i = \exp^{-{{(1-(k-i))}/k}}$ where $i$ stands for the index of the $i^{th}$ previous sample and satisfies $0 \leq i < k$, and $w_i$ is the weight for the $i^{th}$ previous sample. Out of these three filtering strategies, we have used median filtering since it achieves slightly better results. 

\begin{figure}[t!]%
    \centering
    \subfigure[]{%
    \includegraphics[width=0.45\linewidth]{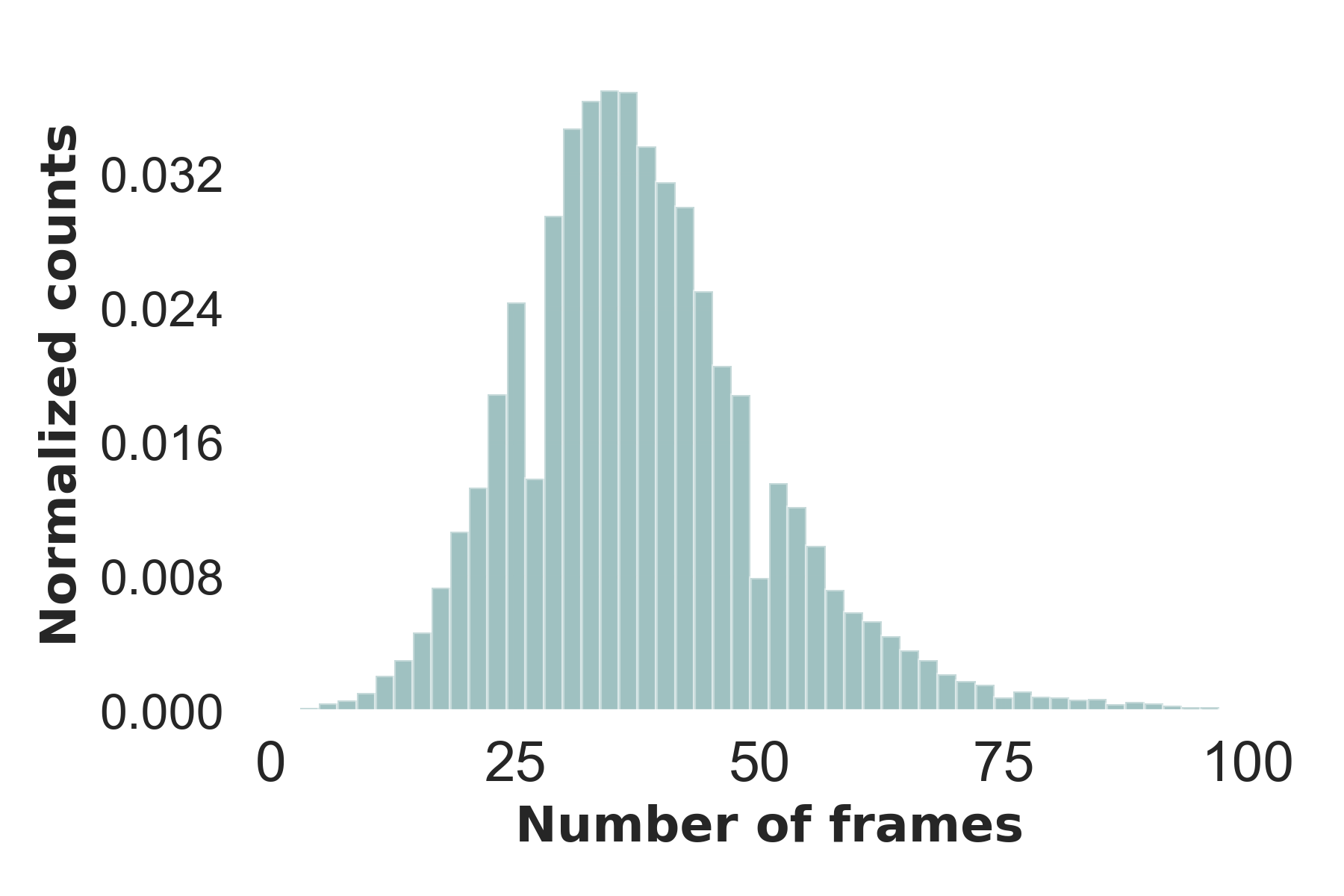}}%
    \label{fig:egoframe}%
    \qquad
    \subfigure[]{%
    \includegraphics[width=0.45\linewidth]{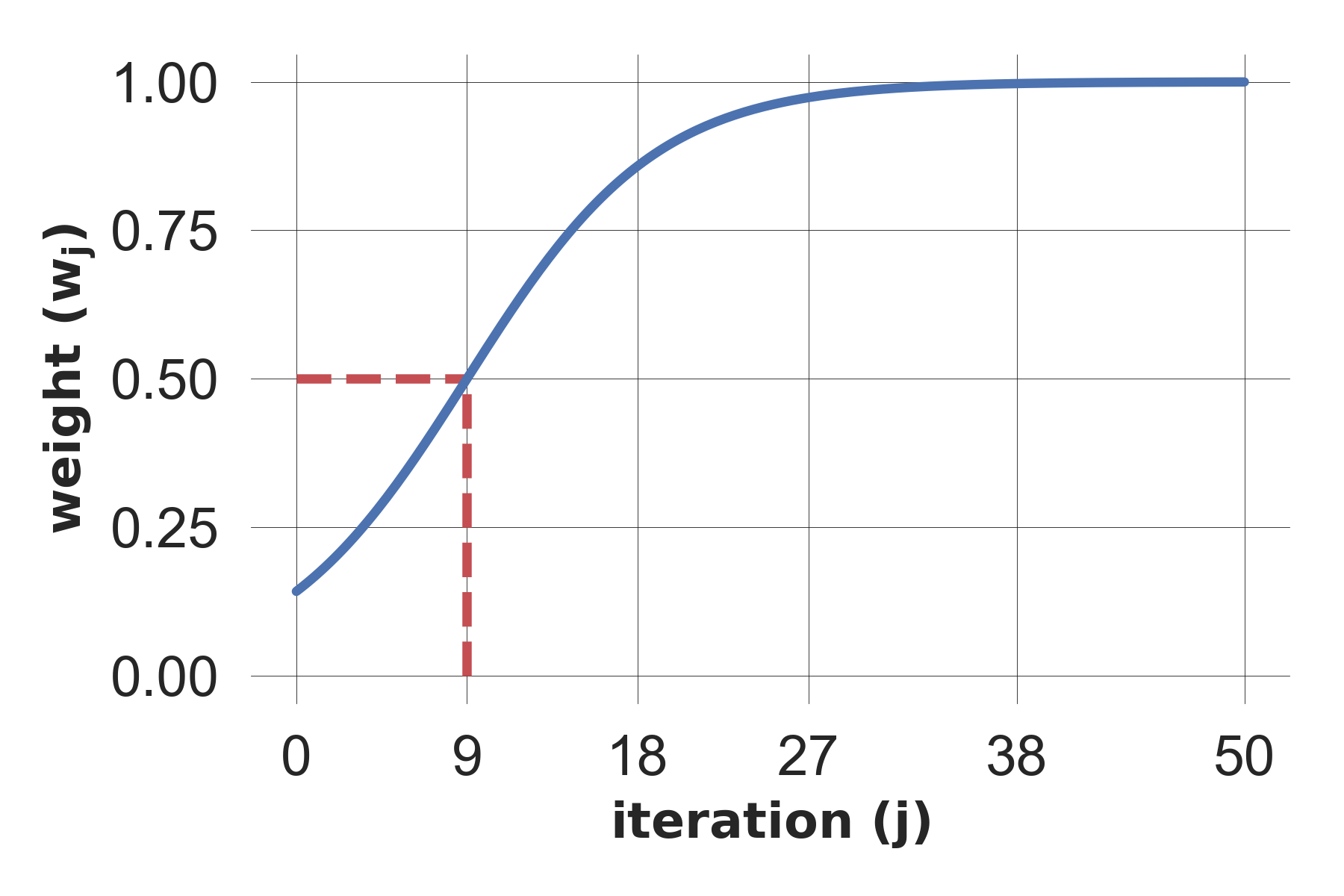}}%
    \caption{(a) Histogram of the gesture durations for the EgoGesture dataset, (b) Sigmoid-like weight function used for single-time activations according to the Equation \eqref{eq:weight}.}
    \label{fig:sigmoid}
\end{figure}

\subsubsection{Single-time Activation}
\label{subsec:sta}

In real-time gesture recognition systems, it is extremely important to have smaller reaction time and single-time activation for each gesture. Pavlovic et al. states that dynamic gestures have \textit{preparation}, \textit{nucleus} (\textit{peak} or \textit{stroke} \cite{mcneill1980conceptual}) and \textit{retraction} parts \cite{pavlovic1997visual}. Out of all parts, nucleus is the most discriminative one, since we can decide which gesture is performed in nucleus part even before it ends.  

Single-time activation is achieved through two level control mechanism. Either a gesture is detected when a confidence measure reaches a threshold level before the gesture actually ends (early-detection), or the gesture is predicted when the detector deactivates the classifier (late-detection). In late-detection, we assume that the detector should not miss any gesture since we assure that the detector has a very high recall rate. 

\begin{figure}[t!]
	\centering
	\includegraphics[width=0.5\textwidth]{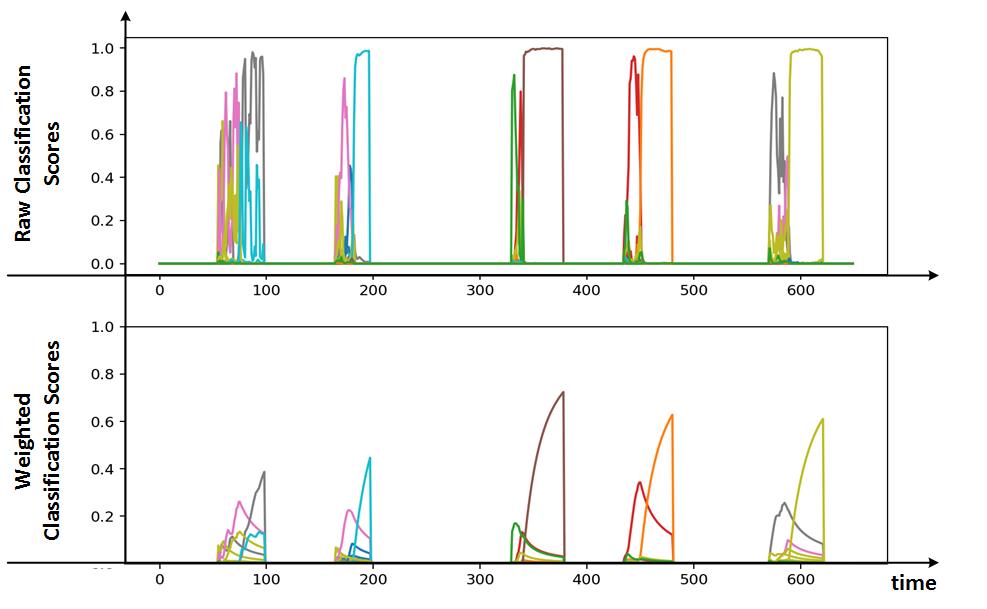}
	\caption{Raw (top) and weighted (bottom) classification scores. At top graph, we observe a lot of noise at the beginning of all gestures; however, near to the end of each gesture the classifier gets more confident. The bottom graph shows that we can remove this noise part by assigning smaller weights to the beginning part of the gestures.}
	\label{fig:weight}
\end{figure}

The most critical part of the early-detection is that, the gestures should be detected after their nucleus parts for a better recognition performance. Because several gestures can contain a similar preparation part which creates an ambiguity at the beginning of the gestures, as can be seen on the top graph of Fig. \ref{fig:weight}. Therefore, we have applied weighted-averaging on class scores with a weight function as in \mbox{Fig. \ref{fig:sigmoid} (b)}, and its formula is given as:

\begin{equation}
    w_j = \frac{1}{(1+\exp^{-0.2\times(j-t)})}
    \label{eq:weight}
\end{equation}
    
\noindent where $j$ is the iteration index of an active state, at which a gesture is detected, and $t$ is calculated by using the following formula:

\begin{equation}
    t = \left \lfloor{\frac{\mu}{4 \times s}} \right \rfloor 
\end{equation}

\noindent where $\mu$ corresponds to the mean duration of the gestures (in number of frames) in the dataset and $s$ is the stride length. For EgoGesture dataset, $\mu$ is equal to 38,4 and for stride of $s = 1$, $t$ is calculated as 9, which is similar for also nvGesture dataset. When a gesture starts, we start to multiply raw class scores with weights $w_j$ and apply averaging. These parameters allow us to have weights equal to or higher than 0.5 in the nucleus part of the gestures on average. Fig. \ref{fig:weight} shows the probability scores of five gestures over each iteration and their corresponding weighted-averages. It can easily be observed that the ambiguity of the classifier at the preparation part of the gestures is successfully resolved with this approach.

\begin{algorithm}[b!]
	\renewcommand{\algorithmicrequire}{\textbf{Input:}}
	\renewcommand{\algorithmicensure}{\textbf{Output:}}
	\caption{Single-time activation in real-time gesture recognition }
    \centering
	\label{online_testing}
	\begin{algorithmic}[1]
		\Require Incoming frames from video data.
		\Ensure Single-time activations.
		\For{each "frame-window" $w_i$ of length $m$}
		\If{a gesture is detected}
		\State{state $\leftarrow$ "Active"}
        \State {$\alpha \leftarrow probs_{j-1}\times(j-1) $}
        \State {$mean_{}probs = (\alpha + w_j\times probs_{j})/j$ }
        \State {$(max_1,max_2)=\max\limits_{gesture}{[mean_{}probs]_2}$ }
        \If{$(max1 - max2) \geq \tau_{\mathbf{early}}$}
        \State {\textit{early-detection} $\leftarrow$ "True"}
        \State \Return {gesture with $max_1$}
        \EndIf
        \State {$j \leftarrow j + 1$}
        \EndIf
        \If{the gesture ends}
        \State{state $\leftarrow$ "Passive"}
        \If{\textit{early-detection} $ \neq$ "True" \& $max_1 \geq \tau_{\mathbf{late}} $}
        \State \Return {gesture with $max_1$}
        \EndIf
        \EndIf
        \State {$i \leftarrow i + 1$}
        \EndFor
	\end{algorithmic}
\end{algorithm}

With this weighted-averaging strategy, we force our single-time activator to make decision at mid-late part of the gestures after capturing their nucleus parts. On the other hand, we need a confidence measure for early-detections in real-time since the duration of gestures varies. Hence, we decided to use the difference between weighted average scores of each classes as our confidence measure for early-detection. When the detector switches the classifier on, weighted average probabilities for each class is calculated at each iteration. If the difference between two highest average probabilities is more than a threshold $\tau_{early}$, then early-detection is triggered; otherwise, we wait for the detector to switch off the classifier and the class with the highest score above $\tau_{late}$ (fixed to 0.15 as it showed the best results in our experiments) is predicted as late-detection. Details for this strategy can be found in Algorithm \ref{online_testing}.

\subsubsection{Evaluation of the Activations}

As opposed to offline testing which usually considers only about class accuracies, we must also consider the following scenarios for our real-time evaluation:

\begin{itemize}
    \item Misclassification of the gesture due to the classifier,
    \item Not detecting the gesture due to the detector,
    \item Multiple detections in a single gesture.
\end{itemize}

Considering these scenarios, we propose to use the Levenshtein distance as our evaluation metric for online experiments. The Levenshtein distance is a metric that measures distance between sequences by counting the number of item-level changes (insertion, deletion, or substitutions) to transform one sequence into the other. For our case, one video and the gestures in this video correspond to a sequence and the items in this sequence, respectively. For example, lets consider the following ground truth and predicted gestures of a video:

\abovedisplayskip=-10pt
\begin{align*}
    & Ground Truth\phantom{aa,,}  [1, 2, 3, 4, 5, 6, 7, 8, 9]  &\\
    & Predicted\phantom{aaaa,}  [1, 2, 7, 4, 5, 6, 6, 7, 8, 9] &
\end{align*}


\begin{table}[t!]
    \centering
    \begin{tabular}{lccc}
        \specialrule{.1em}{.5em}{.5em}
        \multicolumn{1}{c}{\multirow{2}{*}{\textbf{Model}}} & \multicolumn{1}{c}{\multirow{2}{*}{\textbf{Input}}} & \multicolumn{2}{c}{\textbf{Modality}} \\ \cline{3-4} \addlinespace
        \multicolumn{1}{c}{}      & \multicolumn{1}{c}{}    & \textbf{RGB} & \textbf{Depth} \\
        \specialrule{.1em}{.3em}{.3em}
        VGG-16 \cite{zhang_egogesture:_2018}          & 16-frames     & 62.50    & 62.30    \\
        VGG-16 + LSTM \cite{zhang_egogesture:_2018}   & 16-frames     & 74.70    & 77.70    \\
        C3D                                           & 16-frames     & 86.88    & 88.45    \\ 
        ResNeXt-101                                   & 16-frames     & 90.94    & 91.80    \\ 
        C3D+LSTM+RSTTM \cite{zhang_egogesture:_2018}  & 16-frames     & 89.30    & 90.60    \\
        \specialrule{.1em}{.3em}{.3em}
        ResNeXt-101     & 32-frames     & 93.75          & \phantom{\textbf{*}}\textbf{94.03}\textbf{*}             \\ 
        \specialrule{.1em}{.5em}{.5em}
    \end{tabular}
    \caption{Comparison with state-of-the-art on the test set of EgoGesture dataset.}
	\label{tab:egogesture_benchmark}
\end{table}

\begin{table}[t!]
    \centering
    \begin{tabular}{cccc}
        \specialrule{.1em}{.5em}{.5em}
        \multicolumn{1}{c}{\multirow{2}{*}{\textbf{Model}}} & \multicolumn{1}{c}{\multirow{2}{*}{\textbf{Input}}} & \multicolumn{2}{c}{\textbf{Modality}}                                   \\ \cline{3-4} \addlinespace
        \multicolumn{1}{c}{}                       & \multicolumn{1}{c}{}                       & \multicolumn{1}{c}{\textbf{RGB}} & \multicolumn{1}{c}{\textbf{Dept}h} \\
        \specialrule{.1em}{.3em}{.3em}
        C3D             & 16-frames     & 86.88          & 88.45    \\ 
        C3D             & 24-frames     & 89.20          & 89.07    \\ 
        C3D             & 32-frames     & 90.57          & \textbf{91.44 }   \\ 
        \specialrule{.1em}{.3em}{.3em}
        ResNeXt-101     & 16-frames     & 90.94          & 91.80    \\ 
        ResNeXt-101     & 24-frames     & 92.89          & 93.47   \\ 
        ResNeXt-101     & 32-frames     & 93.75          & \phantom{\textbf{*}}\textbf{94.03}\textbf{*}             \\ 
        \specialrule{.1em}{.5em}{.5em}
    \end{tabular}
    \caption{Classifier's classification accuracy scores on the test set of EgoGesture dataset.}
	\label{tab:egogesture_classifier}
\end{table}

For this example, the Levenshtein distance is 2: The deletion of one of "6" which is detected two times, and the substitution of "7" with "3". We average this distance over the number of true target classes. For this case, the average distance is $2/9 = 0.2222$ and we subtract this value from 1 since we want to measure closeness (in this work it is referred as the Levenshtein accuracy) of our results, which is equal to $(1-0.2222) \times 100 = 77.78\%$. 

\section{Experiments}
\label{sec:exp}

The performance of the proposed approach is tested on two publicly available datasets: EgoGesture and NVIDIA Dynamic Hand Gestures dataset. 

\subsection{Offline Results Using EgoGesture Dataset}

EgoGesture dataset is a recent multimodal large scale dataset for egocentric hand gesture recognition \cite{zhang_egogesture:_2018}. This dataset is created not only for segmented gesture classification, but also for gesture detection in continuous data. There are 83 classes of static and dynamic gestures collected from 6 diverse indoor and outdoor scenes. The dataset splits are created by distinct subjects with ratio 3:1:1 resulting in 1239 training, 411 validation and 431 testing videos, having 14416, 4768 and 4977 gesture samples, respectively. All models are first pretrained on Jester dataset \cite{jester}. For test set evaluations, we have used both training and validation set for training.

We initially investigated the performance of C3D and ResNeXt architectures on the offline classification task. Table \ref{tab:egogesture_benchmark} shows the comparison of used architectures with the state-of-the-art approaches. ResNeXt-101 architecture with 32-frames input achieves the best performance.

\begin{table}[t!]
    \centering
    \begin{tabular}{lccc}
        \specialrule{.1em}{0em}{.5em}
        \multicolumn{1}{c}{\multirow{2}{*}{\textbf{Model}}} & \multicolumn{1}{c}{\multirow{2}{*}{\textbf{Input}}} & \multicolumn{2}{c}{\textbf{Modality}}                                   \\ \cline{3-4} \addlinespace
        \multicolumn{1}{c}{}                       & \multicolumn{1}{c}{}                       & \multicolumn{1}{c}{\textbf{RGB}} & \multicolumn{1}{c}{\textbf{Dept}h} \\
        \specialrule{.1em}{.3em}{.3em}
        ResNet-10     & 8-frames      & 96.58          & \phantom{\textbf{*}}99.39\textbf{*}    \\ 
        ResNet-10     & 16-frames     & 97.00          & 99.64    \\ 
        ResNet-10     & 24-frames     & 97.13          & 99.15    \\ 
        ResNet-10     & 32-frames     & 96.65          & \textbf{99.68}    \\ 
        \specialrule{.1em}{.3em}{0em}
    \end{tabular}
    \caption{Detector's binary classification accuracy scores on the test set of EgoGesture dataset.}
	\label{tab:egogesture_detector}
\end{table}

\begin{table}[t!]
	\centering
	\begin{tabular}{lccc}
		\specialrule{.1em}{0em}{.5em}
		\textbf{Modality} & \textbf{Recall} & \textbf{Precision} & \textbf{f1-score}  \\ 
		\specialrule{.1em}{.3em}{.3em}
		RGB            & 96.64	    & 97.10     & 96.87   \\
		Depth          & 99.37		& 99.43		& \textbf{99.40}   \\
		\specialrule{.1em}{.3em}{0em}
	\end{tabular}
	\caption{Detection results of 8-frames ResNet-10 architecture on the test set of EgoGesture dataset.}
	\label{tab:egogesture_det}
\end{table}

Secondly, we investigated the effect of the number of input frames on the gesture detection and classification performance. Results in Table \ref{tab:egogesture_classifier} and Table \ref{tab:egogesture_detector} show that we achieve a better performance as we increase the input size for all the modalities. This depends highly on the characteristics of the used datasets, especially on the average duration of the gestures.

Thirdly, the RGB and depth modalities are investigated for different input sizes. We always observed that the models with depth modality show better performance than the models with RGB. Depth sensor filters out the background motion, and allows the models to focus more on the hand motion, hence more discriminative features can be obtained from depth modality. For real-time evaluation, ResNet-10 with depth modality and input size of 8-frames is chosen as the detector, since smaller window size allows the detector to discover the start and end of the gestures more robustly. The detailed results of this model are shown in Table \ref{tab:egogesture_det}.

\begin{table}[t!]
    \centering
    \begin{tabular}{lcc}
        \specialrule{.1em}{0em}{.5em}
        \multicolumn{1}{c}{\multirow{2}{*}{\textbf{Model}}} & \multicolumn{2}{c}{\textbf{Modality}} \\ \cline{2-3} \addlinespace
        \multicolumn{1}{c}{}    & \textbf{RGB} & \textbf{Depth} \\
        \specialrule{.1em}{.3em}{.3em}
        C3D                                             & 73.86    & 77.18    \\
        R3DCNN  \cite{molchanov2016online}              & 74.10    & 80.30    \\ 
        ResNeXt-101                                     & 78.63    & \textbf{83.82}    \\ 
        \specialrule{.1em}{.3em}{0em}
    \end{tabular}
    \caption{Comparison with state-of-the-art on the test set of nvGesture dataset.}
	\label{tab:nvgesture_benchmark}
\end{table}

\begin{table}[t!]
    \centering
    \begin{tabular}{cccc}
        \specialrule{.1em}{0em}{.5em}
        \multicolumn{1}{c}{\multirow{2}{*}{\textbf{Model}}} & \multicolumn{1}{c}{\multirow{2}{*}{\textbf{Input}}} & \multicolumn{2}{c}{\textbf{Modality}}                                   \\ \cline{3-4} \addlinespace
        \multicolumn{1}{c}{}                       & \multicolumn{1}{c}{}                       & \multicolumn{1}{c}{\textbf{RGB}} & \multicolumn{1}{c}{\textbf{Dept}h}  \\
        \specialrule{.1em}{.3em}{.3em}
        C3D             & 16-frames     & 62.67          & 70.33             \\ 
        C3D             & 24-frames     & 65.35          & 70.33             \\ 
        C3D             & 32-frames     & 73.86          & \textbf{77.18}             \\ 
        \specialrule{.1em}{.3em}{.3em}
        ResNeXt-101     & 16-frames     & 66.40          & 72.82    \\ 
        ResNeXt-101     & 24-frames     & 72.40          & 79.25             \\ 
        ResNeXt-101     & 32-frames     & 78.63          & \phantom{\textbf{*}}\textbf{83.82}\textbf{*}             \\ 
        \specialrule{.1em}{.3em}{0em}
    \end{tabular}
    \caption{Classifier's classification accuracy scores on the test set of nvGesture dataset.}
	\label{tab:nvgesture_classifier}
\end{table}

\subsection{Offline Results Using nvGesture Dataset}

nvGesture dataset contains 25 gesture classes, each intended for human-computer interfaces. The dataset is recorded with multiple sensors and viewpoints at an indoor car simulator. There are in total 1532 weakly-segmented videos (i.e., there are no-gesture parts in the videos), which are split with ratio 7:3 resulting in 1050 training and 482 test videos each containing only one gesture.

We again initially investigated the performance of C3D and ResNeXt architectures on the offline classification task, by comparing them with the state-of-the-art models. As shown in Table \ref{tab:nvgesture_benchmark}, ResNeXt-101 architecture achieves the best performance. Similar to EgoGesture dataset, we achieve a better classification and detection performance as we increase the input size, for all the modalities, as shown in Table \ref{tab:nvgesture_classifier} and Table \ref{tab:nvgesture_detector}. Depth modality again achieves better performance than RGB modality for all input sizes. Moreover, ResNet-10 with depth modality and input size of 8-frames is chosen as the detector in the online testing, whose detailed results are given in Table \ref{tab:nvgesture_det}.

For real-time evaluation, we have selected 8-frames ResNet-10 detectors with depth modality and best performing classifiers in both dataset, which have \textbf{*} sign in corresponding tables.

\subsection{Real-Time Classification Results}

EgoGesture and nvGesture datasets have 431 and 482 videos, respectively in their test sets. We evaluated our proposed architecture on each video separately and calculated an average Levenshtein accuracy at the end. We achieve 91.04\% and 77.39\% Levenshtein accuracies in EgoGesture and nvGesture datasets, respectively.

\begin{figure}[t!]%
    \centering
    \subfigure[]{%
    \includegraphics[height=0.7\linewidth]{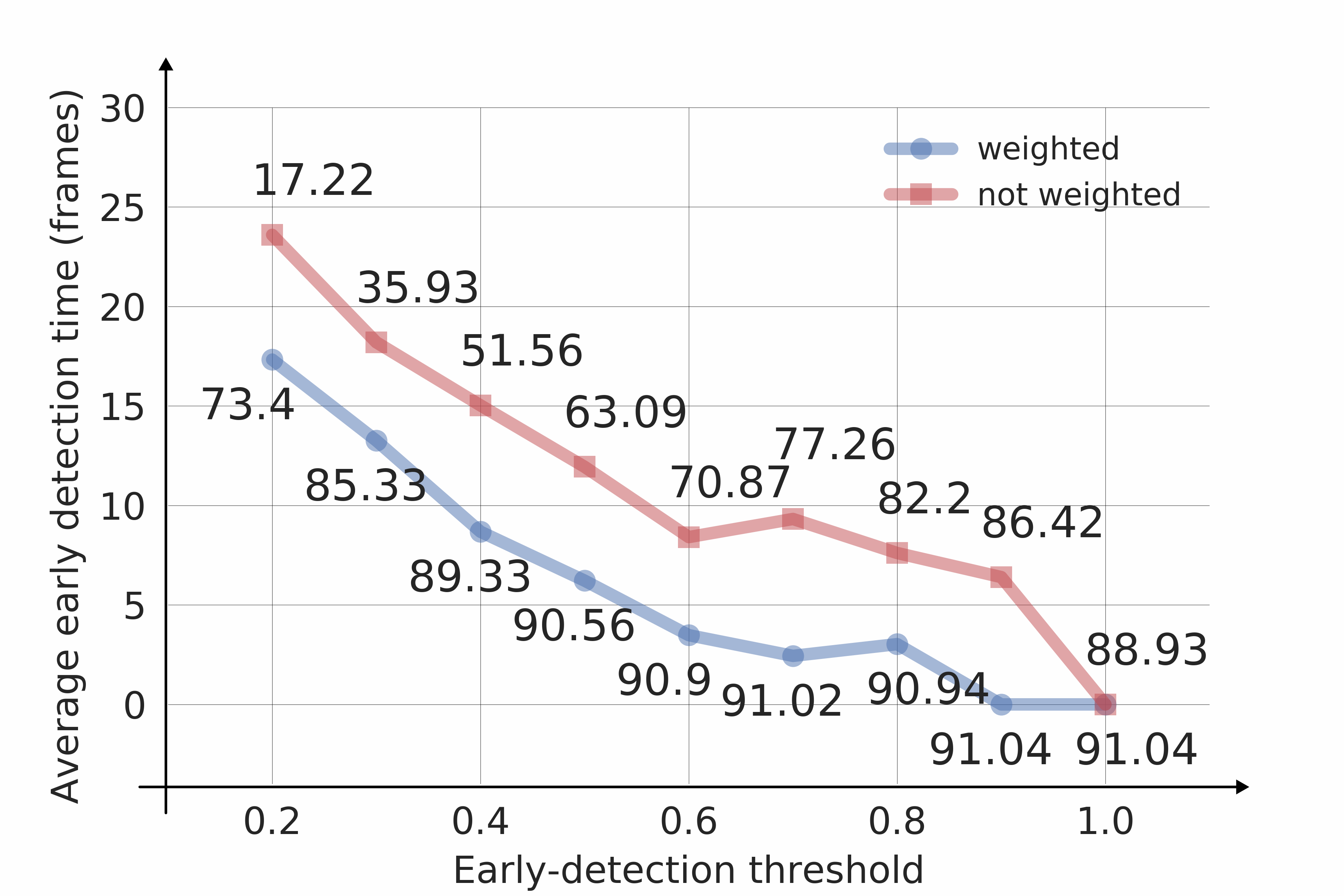}}%
    \label{fig:staego}%
    \qquad
    \subfigure[]{%
    \includegraphics[height=0.7\linewidth]{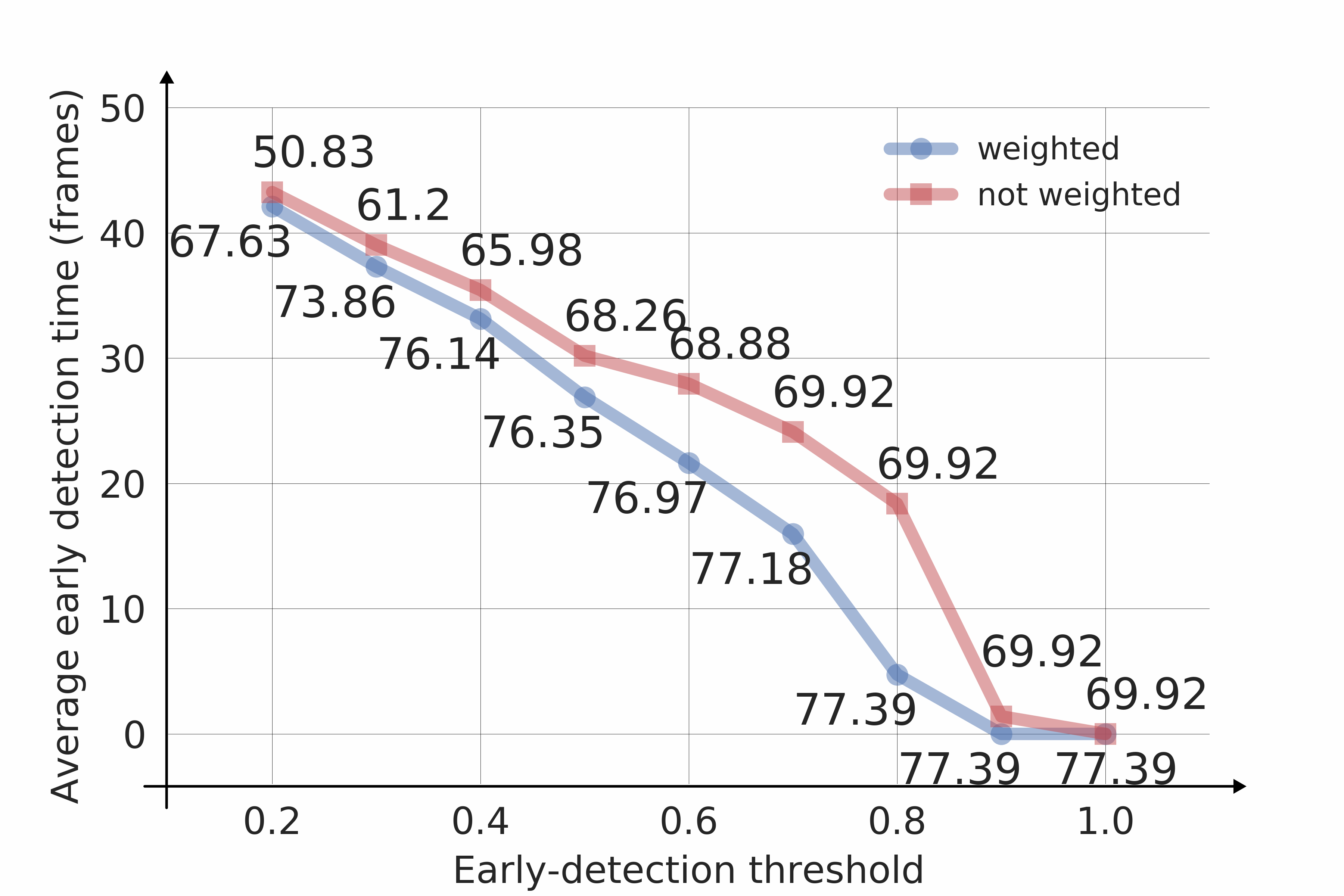}}%
    \caption{Comparison of early detection time, early detection threshold and acquired Levenshtein accuracies for (a) EgoGesture and (b) nvGesture datasets. Numerals on each data point represent the Levenshtein accuracies. Early detection times are calculated only for correctly predicted gestures. Blue color refers to the "weighted" approach in single-time activation, and green color refers to "not weighted" approach. For both datasets, as early detection threshold increases, average early detection times reduce, but we achieve better Levenshtein accuracies.}
    \label{fig:stanv}
\end{figure}

Moreover, the early detection times are investigated by simulating different early-detection threshold levels ($\tau_{early}$) varying from 0.2 to 1.0 with 0.1 steps. Fig. \ref{fig:stanv} compares early detection times of weighted averaging and uniform averaging approaches for both EgoGesture and nvGesture datasets. The Fig. \ref{fig:stanv} shows the importance of weighted averaging, which performs considerably better than uniform averaging. As we increase the threshold, we force the architecture to make decision towards the end of gestures, hence achieving better performance. However, we can gain considerable early detection performance by relinquishing little amount of performance. For example, if we set detection threshold $\tau_{early}$ to 0.4 for EgoGesture dataset, we can make our single time activations 9 frames earlier on average by relinquishing only 1.71\% Levenshtein accuracy. We also observe that mean early detection times are longer for nvGesture dataset since it contains weakly-segmented videos.

\begin{table}[t!]
    \centering
    \begin{tabular}{lccc}
        \specialrule{.1em}{0em}{.5em}
        \multicolumn{1}{c}{\multirow{2}{*}{\textbf{Model}}} & \multicolumn{1}{c}{\multirow{2}{*}{\textbf{Input}}} & \multicolumn{2}{c}{\textbf{Modality}}                                   \\ \cline{3-4} \addlinespace
        \multicolumn{1}{c}{}                       & \multicolumn{1}{c}{}                       & \multicolumn{1}{c}{\textbf{RGB}} & \multicolumn{1}{c}{\textbf{Dept}h}  \\
        \specialrule{.1em}{.3em}{.3em}
        ResNet-10     & 8-frames      & 70.22          &  \phantom{\textbf{}} 97.30\textbf{*}    \\ 
        ResNet-10     & 16-frames     & 85.90          & 97.82    \\ 
        ResNet-10     & 24-frames     & 89.00          & \textbf{98.02}    \\ 
        ResNet-10     & 32-frames     & 93.88         & 97.30    \\ 
        \specialrule{.1em}{.3em}{0em}
    \end{tabular}
    \caption{Detector's binary classification accuracy scores on the test set of nvGesture dataset.}
	\label{tab:nvgesture_detector}
\end{table}

\begin{table}[t!]
	\centering
	\begin{tabular}{lccc}
		\specialrule{.1em}{0em}{.5em}
		\textbf{Modality} & \textbf{Recall} & \textbf{Precision} & \textbf{f1-score}  \\ 
		\specialrule{.1em}{.3em}{.3em}
		RGB            & 70.22	    & 80.31     & 74.93  \\
		Depth          & 97.30		& 97.41		& \textbf{97.35}   \\
		\specialrule{.1em}{.3em}{0em}
	\end{tabular}
	\caption{Detection results of 8-frames ResNet-10 architecture on the test set of nvGesture dataset.}
	\label{tab:nvgesture_det}
\end{table}

Lastly, we investigated the execution performance of our two-model approach. Our system runs on average at 460 fps when there is no gesture (i.e. only detector is active) and 62 (41) fps in the presence of gesture (i.e. both detector and classifier are active) for ResNeXt-101 (C3D) as the classifier on a single NVIDIA Titan Xp GPU with batch size of 8. 

\section{Conclusion}

This paper presents a novel two-model hierarchical architecture for real-time  hand gesture recognition systems. The proposed architecture provides resource efficiency, early detections and single time activations, which are critical for real-time gesture recognition applications. 

The proposed approach is evaluated on two dynamic hand gesture datasets, and achieves similar results for both of them. For real-time evaluation, we have proposed to use a new metric, Levenshtein accuracy, which we believe is a suitable evaluation metric since it can measure misclassifications, multiple detections and missing detections at the same time. Moreover, we have applied weighted-averaging on the class probabilites over time, which improves the overall performance and allows early detection of the gestures at the same time.

We acquired single-time activation per gesture by using difference between highest two average class probabilities as a confidence measure. However, we would like to investigate more on the statistical hypothesis testing for the confidence measure of the single-time activations as a future work. Also, we intend to utilize different weighting approaches in order to increase the performance even further.  
\section*{Acknowledgements}
We gratefully acknowledge the support of NVIDIA Corporation with the donation of the Titan Xp GPU used for this research.

{\small
\bibliographystyle{ieee}
\bibliography{egbib.bib}
}

\end{document}